\documentclass[10pt, a4paper]{article}
\usepackage[utf8]{inputenc} 
\usepackage[T1]{fontenc}    
\usepackage{textcomp}       
\usepackage{tgheros}        

\usepackage{authblk}
\usepackage{caption}
\usepackage{subcaption}
\usepackage{booktabs}
\usepackage{multirow}
\usepackage{threeparttable}
\usepackage{siunitx}
\usepackage[]{lrec2026} 
\usepackage{graphicx}
\usepackage{calc} 
\usepackage{ragged2e} 
\usepackage{tabularx} 
\usepackage{booktabs} 
\usepackage{listings}
\usepackage{listingsutf8} 
\usepackage{newfloat}
\DeclareFloatingEnvironment[
  fileext=lol,
  name=Listing,
  placement={!ht}
]{listing}
\lstdefinelanguage{json}{
  morestring=[b]",
  stringstyle=\color{blue}, 
  morecomment=[l]{//},
  commentstyle=\color{gray}, 
  morekeywords={true,false,null}, 
  keywordstyle=\bfseries\color{teal}, 
  literate=
   *{:}{{{\bfseries:}}}{1}
    {,}{{{\bfseries,}}}{1}
    {[}{{{\bfseries[}}}{1}
    {]}{{{\bfseries]}}}{1}
    {\{}{{{\bfseries\{}}}{1}
    {\}}{{{\bfseries\}}}}{1},
}
\lstdefinestyle{jsonstyle}{
  language=json,
  basicstyle=\ttfamily\footnotesize,
  numbers=left,
  numbersep=6pt,
  stepnumber=1,
  showstringspaces=false,
  breaklines=true,
  breakatwhitespace=false,
  columns=fullflexible,
  keepspaces=true,
  frame=single,
  framerule=0.4pt,
  xleftmargin=1.2em, 
  aboveskip=10pt, 
  belowskip=10pt
}
\usepackage{caption}
\newcommand{\RowHeight}{0.28\textheight} 
\newcommand{\HGap}{1em} 
\newcommand{\VGap}{0.75em} 
\newlength{\TwoColW}
\title{If I Could Turn Back Time:\\ Temporal Reframing as a Historical Reasoning Task for LLMs}

\author{
Lars Bungum\textsuperscript{1} \and Charles Yijia Huang\textsuperscript{1} \and Abeer
Kashar\textsuperscript{2} \\
\textsuperscript{1} NTNU, Norway \\
\texttt{lars.bungum@ntnu.no, charleyh@ntnu.no} \\
\textsuperscript{2}University of Waterloo,  Canada \\
\texttt{akashar@uwaterloo.ca}
}

\abstract{
In this study, we experiment with the ability of LLMs to do temporal reasoning.  Using a Norwegian book from 1940 containing trivia questions, we prompt the LLMs to answer the questions as if it were 1940.  We also pose the questions in both English and Norwegian.  Correct answers are often presented as sentences, and grading is done by means of LLM-as-judge, with sampled checks by a native speaker.  Prompting in English consistently gave better results than in Norwegian, an unexpected result.  In contrast, using larger LLMs improved results.  We tested the DeepSeek-R1, Gemma3, Qwen3, and Llama3.1 model families, and also the largest available LLM especially crafted for Norwegian.
 \\ \newline  }

\begin{document}
\maketitle

\section{Introduction}
\label{sec:introduction}

Temporal Reframing (TR) is a challenging cognitive task for humans and machines alike, as what was correct in 1940, may no longer be in 2025. In this work, we use a quiz book from 1940, "Vet De Det", pseudonymously published by Hugin Ravn, henceforth called "The Book". It is used to measure how well Large Language Models (LLMs) are able to answer questions as temporally reframed to the year 1940. The Book consisted of 1000 questions in 25 categories in Norwegian, whose labels were sometimes indicative of their content (\emph{Around the World}), and other times not (\emph{First Round}).

To illustrate how challenging the task is, consider the question \emph{"What country has an area that exceeds that of the United States of America and Canada together?"}, which in 1940 was true for the Soviet Union (~22Mkm²>~19Mkm²), but not for Russia (~17Mkm²). 
As such, the task of pretending that it is some arbitrary time period, for the model, requires what for human participants would be described as reasoning. Inasmuch as the LLMs mimic such behavior, this also requires reasoning on their part. Without full knowledge of how openly available, but commercial models are built, we consider it unlikely that they have been explicitly trained for TR, and even more unlikely for the year 1940.

Logical tasks are often divided into a) inductive, b) deductive, or c) abductive reasoning tasks \cite{XU2025101370}. They denote reasoning tasks that identify patterns from observations, reach conclusions based on premises with certainty, and forming the most likely explanation without certainty, respectively.  The task at hand is primarily deductive, but also employs some abductive reasoning  for cases where relevant information if internal facts in the model are incomplete. Such a condition could arise for some questions in The Book.

\citet{cao2025generalizableevaluationllmera} note that it is hard to design datasets that do not simply measure Norwegian proficiency or shallow pattern matching. Regarding the discussion of whether LLMs at all reason (or think), we consider in the following that LLMs are reasoning when they simulate behavior that would require reasoning on the part of humans, despite that they are constructed from simpler building blocks without explicit logical rules.

Due to the historical reasoning element, TR stands out from traditional Question-Answering tasks, which use LLMs as knowledge bases.  For each question, the LLMs must consider the premise that questions should be answered as if we were in 1940.  While LLMs could be trained on a dataset as our 1940 instance, it is inconceivable that such instruction-datasets could exist for all periods, as questions could be posed for events from geological time and beyond. Thus, what we define as reasoning abilities is necessary to find the right answer.

Our study investigates two main Research Questions for both international and Norwegian models: 
\begin{itemize}
    \item \textbf{RQ1}: How does performance scale with parameter size?
    \item \textbf{RQ2}: What is the effect of prompting in English and Norwegian?
\end{itemize}

We conducted experiments on the DeepSeek-R1 \cite{deepseekai2025deepseekr1incentivizingreasoningcapability}, Gemini \cite{geminiteam2025geminifamilyhighlycapable}, Llama3.1\cite{grattafiori2024llama3herdmodels}, and Qwen3 \cite{yang2025qwen3technicalreport} LLM families for the scaling experiments, as well as running the experiments on three different models developed especially for Norwegian. Those were two models from the Mimir projet \cite{delarosa2025impactcopyrightedmateriallarge}, and the recently published NorwAI-Magistral24B-Reasoning (henceforth: NorwAI-Magistral24B) model\footnote{\url{https://huggingface.co/NorwAI/NorwAI-Magistral-24B-reasoning}}.

Questions are always posed in Norwegian, but the prompts are also presented in English for comparison. The prompts used both for  querying models and grading are presented in Appendix~\ref{appendix}.

\subsection{Contributions}

This paper makes the following contributions:
\begin{itemize}
    \item We cast Temporal Reframing as a Historical Reasoning task as a novel task for Norwegian.
    \item We investigate the scaling effects on four families of LLMs.
    \item We successfully apply LLM-as-judge scoring with a thorough qualitative error analysis.
    \item We apply the recent  NorwAI-Magistral24B model to the problem, the first Norwegian model of its kind.
    \item We find that the problem is not trivial even for the largest open models, and make an error analysis looking into why.
    \item We lay the foundation for future research.
\end{itemize}

\section{Related Work}
\label{sec:related:work}
This work falls within the scope of overall evaluation of LLMs, but more specifically relates to the
specific evaluation of temporal reasoning capabilities.  Evaluation of LLMs is a broad topic, which
is important for academia and industry actors alike. As an example, DeepSeek-R1 models are presented
on their Huggingface\footnote{\url{https://huggingface.co/deepseek-ai/DeepSeek-R1}} page with their
performance on the datasets AIME
2024\footnote{\url{https://artofproblemsolving.com/wiki/index.php/2024_AIME_I_Problems}},
Codeforces\footnote{\url{https://codeforces.com}}, GPQA-Diamond \cite{rein2024gpqa}, Math-500
\cite{lightman2024verify}, MMLU \cite{hendrycks2021measuring}, and
SWE-Bench\cite{jimenez2024swebench} verified.  Additionally, many independent \emph{leaderboards}
exist, which combine performance on many evaluation dimensions and present an overall ranking.
Examples are the \emph{Open Large Norwegian Model
Leaderboard}\footnote{\url{https://huggingface.co/organizations/open-llm-leaderboard/}} at
Huggingface, or the \emph{Norwegian Model Evaluation Harness} by EleutherAI \cite{eval-harness}, or
the Euroeval\footnote{Formerly known as ScandEval.} framework \cite{nielsen2023scandeval}.  The
leaderboards focus on different tasks, and subdivisions thereof.  Euroeval, for instance, divides
subtasks into Natural Language Generation and Understanding.

LLM evaluation has been surveyed extensively, e.g., in the influential but dated survey by \citet{chang2023surveyevaluationlargelanguage}, which analyzed what to evaluate, where to evaluate it, and how to evaluate it.  Also, the above-mentioned survey by \citet{cao2025generalizableevaluationllmera} aims to go beyond benchmarks to core capabilities they exemplify as knowledge, reasoning, instruction following, multi-modal understanding, and safety, as well as highlighting LLM-as-judge approaches to answer the need for automated evaluation.  \citet{peng2024surveyusefulllmevaluation} looked beyond core capabilities and argue that LLMs have to be evaluated as agents to be useful.

With regards to temporal reasoning, there has been work on incorporating temporal knowledge into LLMs.  \citet{wang2025effectivetimeawarelanguagerepresentation} built the BiTimeBERT model, which is a time-aware masked language modeling, and citing promising results on tasks like Named Entity Disambiguation, for which reasoning in time could be useful.  \citet{uddin2025unseentimeqatimesensitivequestionansweringllms} created the UnSeenTimeQA dataset, a Time Sensitive Question Answering (TSQA) benchmark.  It is comprised of questions of what some person was doing in a given year, or reasoning about time intervals.  The authors are concerned with the abilities of present datasets to measure temporal reasoning as opposed to recalling memorized knowledge like \emph{"Where did Diego Maradona play in 1987?"}, and address this by avoiding web-searchable questions.  Other TR datasets exist, such as Time-Sensitive-QA \cite{chen2021datasetansweringtimesensitivequestions} and TempReason \cite{tan2023benchmarkingimprovingtemporalreasoning} that were based on mining Wikipedia for questions. \citeauthor{tan2023benchmarkingimprovingtemporalreasoning} outline three levels of temporal reasoning; time-time relations (\emph{What is the year after 2010?}), time-event relations (as the Maradona question above), and event-event reasoning (\emph{What team did Pelé play for after Santos?)} Published in the early stages of the Generative AI avalanche,  these datasets were vulnerable to data leakage. 

More recently, \citet{pham2025sealqaraisingbarreasoning} launched SealQA, a benchmark based on constantly mining fresh questions, that would be able to test the reasoning capabilities of LLMs when the answer to the questions were given as context, and could not be found in the training material.  Also, \citet{wei2025time}\footnote{Accepted to NeurIPS according to the authors' website, last visited on October 24, 2025.} launched the Time dataset, consisting of 38,522 QA pairs across eleven subtasks spanning three levels: basic temporal understanding (e.g., \emph{"What happened first?"}), temporal expression reasoning (e.g., \emph{"What happened between two dates in 2011?"}), and complex temporal relationship reasoning, including counterfactual and timeline ordering tasks.  A counterfactual time reasoning question could be asking how old the Berlin Wall would have been had it fallen 8 years later.
  
Our approach, with 1000 questions with a cutoff year of 1940, does not test the breadth of TR as in the related work, but offers an in-depth insight into how well LLMs do on time-travel to 1940, without being provided any other context.  As such, all of the questions are time-event reasoning, and some require event-event reasoning, as the question on land area (see Section~\ref{sec:introduction}).  Furthermore, our foray into TR for Norwegian offers a narrow, but deep account of state-of-the-art models' performance on questions from 1940. To the best of our knowledge, no TR research has been published for Norwegian.

\section{Method}
\label{sec:method}
The method can be described in three steps.  1) creating a dataset from the 1940 book, 2) querying LLMs on these questions, and 3) evaluating the questions with a different LLM.

\subsection{Dataset Creation}
\begin{listing*}[t]
\caption{Example JSON payload for the example question in Section~\ref{sec:introduction}.}\label{lst:json}
\label{listing:json}
\begin{lstlisting}[style=jsonstyle]
      "qa_pairs": [
    {
      "question": "Hvilken fastlandsstat er stoerre enn U.S.A. og Canada tilsammen?",
      "answer": "Sovjet-Samveldet (21.6 mill. km(*@\textsuperscript{2}@*), 15 % av jordoverflaten)."
    },
\end{lstlisting}
\end{listing*}

The Book was scanned and read with Optical Character Recognition (OCR).  The structure of the book was to have questions in the first part and answers in the back, which meant that questions and answers needed to be paired in the resulting JSON file.  The resulting text was also manually checked for errors in OCR.  The Python package \texttt{pytesseract}\footnote{\url{https://github.com/madmaze/pytesseract}} was used for the OCR, and a custom script was created to connect questions with answers to produce the final JSON output exemplified in Listing~\ref{listing:json}.

\subsection{Querying Models}

In the second phase, Ollama\footnote{\url{https://ollama.com/}} servers  were instantiated on the HPC cluster IDUN \cite{sjalander+:2019epic} for interaction with the models.  With the exception of the largest DeepSeek-R1 model, all models could fit on one GPU.  The largest model required four H200 GPUs, and Ollama conveniently handles the distribution of the model to more GPUs.  

The models were provided with a system prompt, either in English and Norwegian in separate instances, instructing them to only apply knowledge prior to 1940 in their answers. Subsequently, the models were posed the  questions from the dataset in the user prompt separately.  The output was stored to file. 

\subsection{LLM-as-judge Evaluation}

Because of the nature of the dataset, where answers were provided as explanations, grading the LLM output was a challenge.  With separate experiments for two languages for 20 models, each generating 1000 answers, some automatic solution was required.  We considered converting the dataset to a multiple choice format, but opted for an LLM-as-judge \cite{li2024llmsasjudgescomprehensivesurveyllmbased} solution. 

 We appointed gpt-oss:120B \cite{openai2025gptoss120bgptoss20bmodel} as judge.  This is a strong model, but not part of the LLM families we would select for experiments (see below).  The model was presented with the suggested answer from an LLM agent and the correct answer from the book.  Based on this answer, it was instructed to determine whether the answers were true or false using one-shot prompting. 

\begin{table*}[t]
  \centering
  \caption{Manually identified grading mistakes in the questions considered wrong for DeepSeek-R1:671B. }
  \label{tab:wrong_grading}

  \setlength{\tabcolsep}{5pt}        
  \renewcommand{\arraystretch}{1.15} 
  \footnotesize                      

  \begin{tabularx}{\textwidth}{
      >{\RaggedRight\arraybackslash}p{.34\textwidth}
      >{\RaggedRight\arraybackslash}p{.24\textwidth}
      >{\RaggedRight\arraybackslash}X
  }
    \toprule
    \textbf{Question (in English)} &
    \textbf{Answer (abbreviated)}  &
    \textbf{Explanation} \\
    \midrule

    What are the largest marsupials? &
    Kangaroos.  &
    The correct answer was Kangaroos (in Australia).  The grader could have given it an F because it failed to mention the country. \\

    Who invented the execution machinery used during the French revolution? &
    It was requested by Dr. Guillotin, but designed by Dr. Louis and prototyped by Tobias Schmidt. &
    Since Guillotin was mentioned first, and the others named thereafter, this should, arguably count as correct.  \\

    How long has England (sic) owned Gibraltar? &
    227 years. &
    The correct answer from The Book was 250 years, but 1940-1704=236, so the model's answer was actually closer. It is unclear whether this is due to the "wrong" correct answer in The Book, or the grader's inability to see how close the answer was to the correct. \\

    What was the full name of Charles XIV John? &
    Jean-Baptiste Jules Bernadotte. &
    The correct answer was Jean Baptiste Bernadotte (1763-1844) and the answer should clearly have been graded as correct. \\
    
    What is the population of the United States? &
    123 million. &
    The correct answer is given as 130 million, which is close enough.  It is unclear whether the
grader gets it wrong due to superfluous and erroneous reasoning in the candidate answer, or if it
does understand how close the numbers are.  \\

    What does the hedgehog mostly eat? &
    Insects, beetles, and worms. It can also eat frogs, toads, and rodents. &
    The correct answer was insects and sometimes worm and snails, which is very close, and should
have been graded as correct.  Possibly, the mention of rodents made the grader too strict.  \\

    What are the "headhunters" on Borneo called? &
    Dayaks, including the Iban, Bidayuh and Kayan peoples. &
    The correct answer was Dayaks with a different spelling (\emph{"Dajaker"}).  The extra
information may have nudged the grader to a false verdict.  \\

    What job did Alexander Kielland have when he died? &
    Amtmann i Romsdals Amt.  &
    The correct answer was given as Amtmann in Møre, which is incorrect.  The Amt was actually
called Romsdals Amt. However, the model's answer incorrectly added that he had the position from
1891 (1902 was correct), which could have confused the grader further. As a consequence, this is
more of a dataset error than a grader error.  \\

    What living Russian emigrant author received the Nobel Prize? &
    Ivan Bunin. &
    The correct answer was given as "Ivar Bunin". This was again a dataset error, more than a grader
error.  \\
    
    \bottomrule
  \end{tabularx}
\end{table*}
 
Table~\ref{tab:wrong_grading}\footnote{Translations are provided by one of the authors who is a native speaker. Norwegian originals are omitted for brevity.} shows nine answers that would likely have been graded as correct by a human, despite some superfluous information.  Two of the mistakes were due to mistakes in the correct answer section of The Book.  This amounts to 1\% of the questions, which means that the reported scores should have been a percentage point higher for the strongest model.  For the correct answers, samples were inspected and no mistakes were found. As a consequence, we assume that such errors are less frequent. 
 
\subsection{Model Selection}

For our experiments, we selected two groups of Model families, henceforth, Groups A and B.  The models in Group A were chosen because they were offered multiple data points, that they were easily available through Ollama's library, and also that they are widely known and recognized as well-performing on reasoning tasks.  As mentioned in Section~\ref{sec:introduction}, the model families were DeepSeek-R1,Gemma3, Llama3.1, and Qwen3.

All models were quantized with the default Q4\_K\_M, except for the NorwAI-Maigstral24B model, which was downloaded directly from Huggingface and quantized with Q8\_K\_M. In addition to this model, we experimented with two smaller Mimir models. The difference between the Scratch and Core models that the latter use pretrained weights from Mistral 7B v0.1 \cite{delarosa2025impactcopyrightedmateriallarge}.
Finally, we ran the same experiments on the two available versions of the gpt-oss  model, mostly out of curiosity of how it would perform on a task it graded itself.  Altogether these four models comprised Group B.

\begin{figure*}[h]

  \centering
  \caption{Group A Model Performance}
  \label{fig:fullpage}

  \setlength{\tabcolsep}{0pt}
  \renewcommand{\arraystretch}{1}

  \begin{tabular}{p{.5\textwidth}@{\hspace{\HGap}}p{.5\textwidth}}
    \subcaptionbox{DeepSeek-R1 Model Performance\label{fig:a}}{%
      \includegraphics[width=\linewidth,height=\RowHeight,keepaspectratio]
        {\detokenize{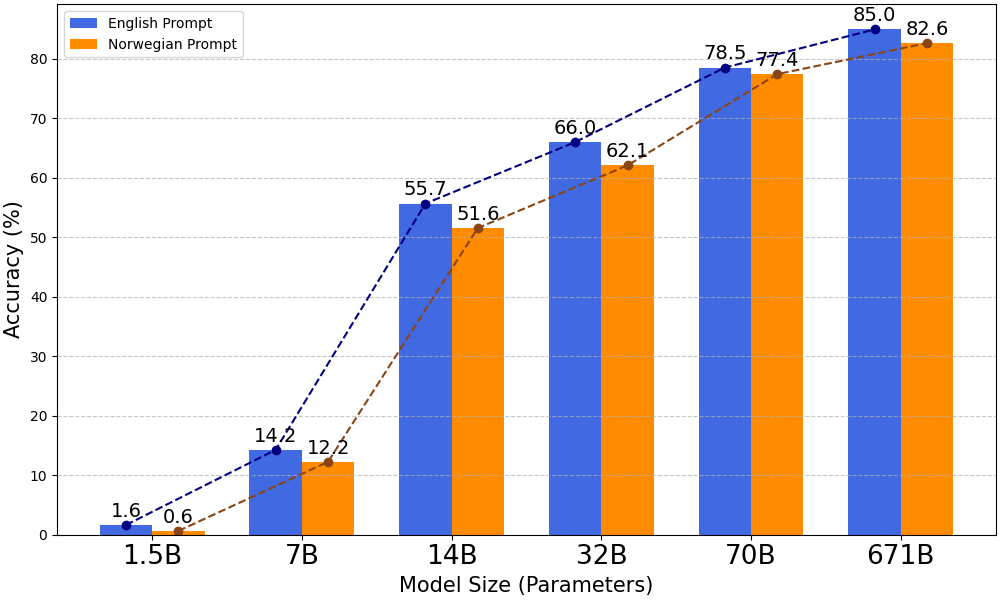}}} &
    \subcaptionbox{Gemma3 Model Performance\label{fig:b}}{%
      \includegraphics[width=\linewidth,height=\RowHeight,keepaspectratio]
        {\detokenize{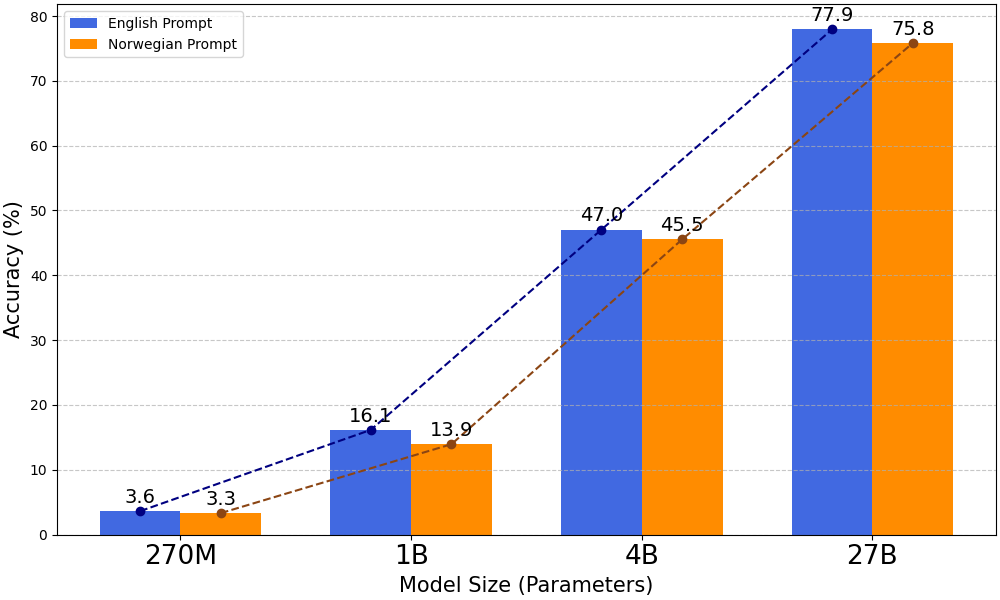}}}
    \\
    [\VGap]
    \subcaptionbox{Qwen3 Model Performance\label{fig:c}}{%
      \includegraphics[width=\linewidth,height=\RowHeight,keepaspectratio]
        {\detokenize{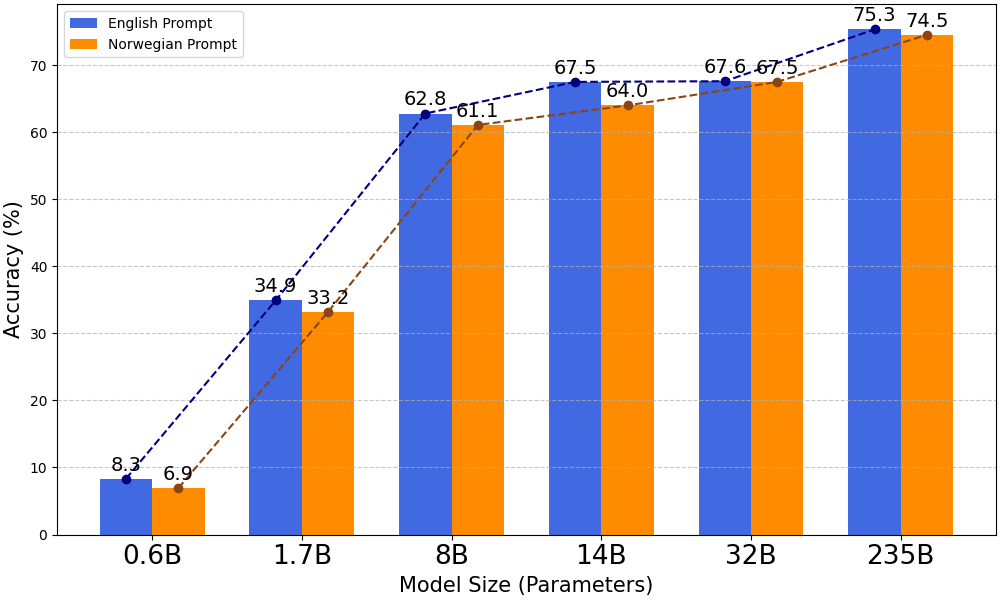}}} &
    \subcaptionbox{Llama3.1 Model Performance\label{fig:d}}{%
      \includegraphics[width=\linewidth,height=\RowHeight,keepaspectratio]
        {\detokenize{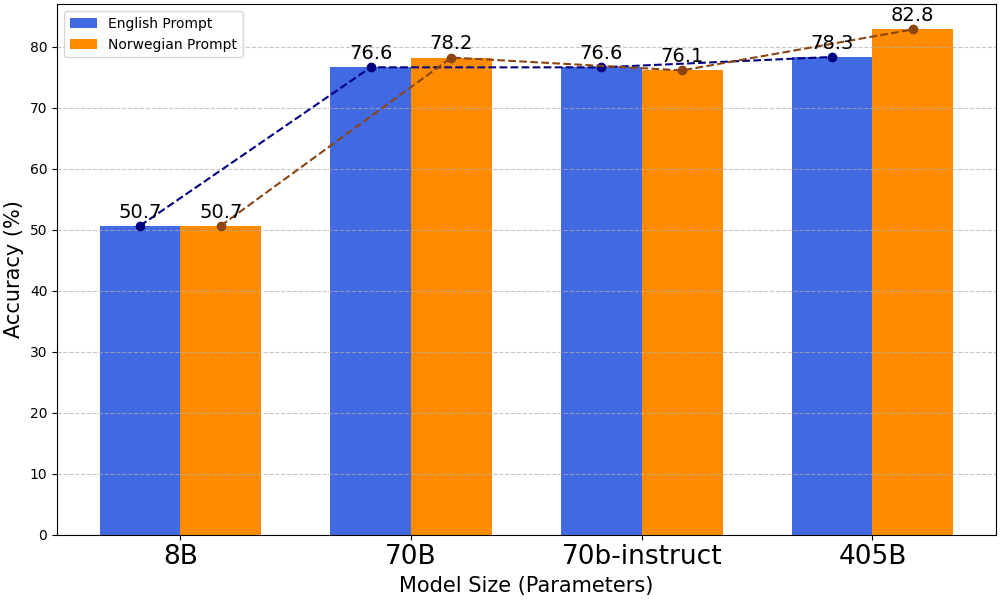}}}
  \end{tabular}
\end{figure*}

\begin{table}[h]
\centering
\tiny
\setlength{\tabcolsep}{6pt}
\renewcommand{\arraystretch}{1.15}
\begin{threeparttable}
\caption{Accuracy (\%) – Model Group A.}
\label{tab:accuracy-group-a}
\begin{tabular}{ll
                S[table-format=3.1]
                S[table-format=3.1]}
\toprule
\textbf{Family} & \textbf{Model/Size} & {\textbf{EN (\%)}} & {\textbf{NO (\%)}} \\
\midrule
\multirow{6}{*}{DeepSeek‑R1}
  & 671B & 85.0 & 82.7 \\
  & 70B  & 78.6 & 77.5 \\
  & 32B  & 66.1 & 62.2 \\
  & 14B  & 55.8 & 51.7 \\
  & 7B   & 14.2 & 12.2 \\
  & 1.5B &  1.6 &  0.6 \\
\midrule
\multirow{4}{*}{LLaMA 3.1}
  & 405B                  & 78.3 & 82.8 \\
  & 70B                   & 76.6 & 78.2 \\
  & 70B-Instruct & 76.6 & 76.1 \\
  & 8B                    & 50.7 & 50.7 \\
\midrule
\multirow{4}{*}{Gemma3}
  & 27B  & 78.0 & 75.9 \\
  & 4B   & 47.2 & 45.7 \\
  & 1B   & 16.2 & 14.1 \\
  & 270M &  3.6 &  3.3 \\
\midrule
\multirow{6}{*}{Qwen3}
  & 235B & 75.3 & 74.5 \\
  & 32B  & 67.7 & 67.6 \\
  & 14B  & 67.6 & 64.1 \\
  & 8B   & 62.8 & 61.1 \\
  & 1.7B & 35.0 & 33.3 \\
  & 0.6B &  8.3 &  6.9 \\
\bottomrule
\end{tabular}
\end{threeparttable}
\end{table}

\begin{table}[h]
\centering
\tiny
\setlength{\tabcolsep}{6pt}
\renewcommand{\arraystretch}{1.15}
\begin{threeparttable}
\caption{Accuracy (\%) - Model Group B.}
\label{tab:accuracy-group-b}
\begin{tabular}{ll
                S[table-format=3.1]
                S[table-format=3.1]}
\toprule
\textbf{Family} & \textbf{Model/Size} & {\textbf{EN (\%)}} & {\textbf{NO (\%)}} \\
\midrule
\multirow{2}{*}{gpt-oss}
  & 120B & 80.6 & 79.0 \\
  & 20B  & 66.7 & 65.6 \\
\midrule
NorwAI-Magistral24B (Q8\_0)
  & 24B & 75.5 & 76.2 \\
\midrule
\multirow{2}{*}{Mimir-Mistral}
  & 7B-Scratch‑Instruct &  7.9 &  6.5 \\
  & 7B-Core‑Instruct    & 31.1 & 29.5 \\
\bottomrule
\end{tabular}
\end{threeparttable}
\end{table}

\section{Experiments and Results}

All our experiments follow the same trajectory as described in Section~\ref{sec:method}.  For each LLM, we pose the questions in English and Norwegian, to be graded by our evaluation script.  The Group A experiments were designed to investigate the scaling effect on TR and Group B was to sample the performance on Norwegian-focused LLMs primarily, with the less relevant performance on the grader model as an addition.

Figure~\ref{fig:fullpage} gives an overview of the LLMs in Group A.  Subfigures a and c have six (twelve) data points, which could seem to indicate fitted curves with a maximum not far after the largest models.  Especially for the DeepSeek-R1 models, the incline gradually reduces (after 7B).  However, one should be careful with this interpretation, as some qualities can appear in LLMs only after an unknown parameter size.  Such emergent properties are implied to come without specific training  and appear as a qualitative changes in behavior emerge as a consequence of quantitative changes in the system \cite{lu-etal-2024-emergent}. While there could be further jumps in performance, the Figure clearly demonstrates a clear trend between model size and performance the TR task. 

Tables~\ref{tab:accuracy-group-a}~and~\ref{tab:accuracy-group-b} complete the picture.  For Group B, a graphical representation would have offered little and was omitted.  It is worth noting that the NorwAI-Magistral24B model was quantized with 8 bits per weight and the rest with 4. On the other hand, the parameter size falls between the data points in the other LLM families.  However, the limited performance for the 7B models (for Mimir-7B-Scratch-Instruct and DeepSeek-R1 especially, the effect was not so strong for Qwen3 and Gemma3) could mean that the TR capability appears only beyond this size for some models.

With regards to our second RQ on prompting language, we see that the English prompts are consistently better for all LLM families with the exception of Llama3.1, where the performance is on par until the largest 405B model, where Norwegian prompts are doing better by a margin of four percentage points.  It is also the case for the NorwAI-Magistral24B model's only data point, but not for the Mimir-Mistral models of smaller size.  All experiments were run only once out of concern for resource use.  Since the Llama3.1:405B model was an anomaly with regard to prompting language, however, the experiment was run twice and the average of the two runs are presented in the results to corroborate the finding.

\subsection{Error Analysis}

\begin{table*}[t]
  \centering
  \caption{Examples of wrong answers from the 21 answers all models got wrong with example answers by DeepSeek-R1:671B. }
  \label{tab:wrong_answers}

  \setlength{\tabcolsep}{5pt}        
  \renewcommand{\arraystretch}{1.15} 
  \footnotesize                      

  \begin{tabularx}{\textwidth}{
      >{\RaggedRight\arraybackslash}p{.34\textwidth}
      >{\RaggedRight\arraybackslash}p{.24\textwidth}
      >{\RaggedRight\arraybackslash}X
  }
    \toprule
    \textbf{Question (in English)} &
    \textbf{Answer (abbreviated)}  &
    \textbf{Explanation} \\
    \midrule

    Where does the term “\emph{lotter}” come from? &
    The Norse word “\emph{lotr}”, meaning a slob. &
    The term refers to Lotta Svärd from \emph{The Tales of Ensign Stål} by J.\,L. Runeberg, and the model confuses its origin with another word. \\

    What is the biggest name in Norwegian speed walking? &
    Sonja Henie &
    An old term for the sport was used, \emph{gangsport}, and the model answered with the greatest figure skater in Norwegian history. \\

    What was the name of the last Swedish “\emph{stattholder}” (top civil servant) in Norway? &
    Severin Løvenskiold &
    While Løvenskiold was “\emph{stattholder}”, the model misses that he was Norwegian. \\

    What was the popular name of Christian~V’s Norwegian Law of~1687? &
    Kong Christian den Femtis Norske Lov &
    The model answers with the correct law (albeit misspelled) but misses the moniker “\emph{Kristian Kvint}”. \\

    What is Norway’s strongest lighthouse? &
    Strongman Jo Visdal (1871–1923). &
    “\emph{Fyr}” (lighthouse) is homonymous with the word for “guy”. Especially poor answer because there was a sculptor Jo Visdal (1861–1923). \\
    What is Norway's largest dam? &
    Lake Mjøsa. &
     The model answers the largest (natural) lake in Norway. \\
    \bottomrule
  \end{tabularx}
\end{table*}

21 questions were answered incorrectly by all models.  We considered them the most difficult questions, and extracted them for qualitative evaluation and we use the answer provided by the strongest model, DeepSeek-R1:671B to illustrate what went wrong.  Table~\ref{tab:wrong_answers}\footnote{Translations are provided by one of the authors who is a native speaker. Norwegian originals are omitted for brevity.} shows a selection of wrong answers with an explanation of what went wrong. After perusal, it is clear that the models struggle with ambiguous words where the intended meaning is no longer used frequently or obsolete terms.  This is true for the speed walking term, "\emph{gangsport}", "\emph{Kristian Kvint}", who now is usually referred to as Christian V, and also the word for "dam", which in present Norwegian usually references an artificial lake, but more rarely a reservoir.  Otherwise, it is clear that the models confidently provide answers even when misunderstanding the questions completely.  Such unsubstantiated answers are often termed hallucinations \cite{Farquhar2024SemanticEntropy}.

\section{Discussion}
\label{sec:discussion}

Temporal Reframing poses a very difficult cognitive problem. As the discussion of related work in Section~\ref{sec:related:work} reflected, it can be difficult to separate fact retrieval from reasoning in the behavior exhibited by the LLMs.  Table~\ref{tab:wrong_answers} reflects that the models struggle with ambiguous words, especially when there have been semantic shifts for terms over time.  In addition to this list, the strongest DeepSeek-R1 model gave an insightful incorrect answer for the following question: \emph{What were the names of the most famous writers in German literature at the start of the last century?}, which it answers with Rainer Maria Rilke and Stefan George, who were writers active in the beginning of the 20th century.  This means that the model could not reason that it had to pick the most famous writers from at the start of the 19th century, where Goethe and Schiller should have been easy answers.  

Overall, the performance of the strongest models is very good, and we expect that it would be better than many, if not most, Norwegians living today.  However, the performance is not stellar, and it is not solved, even for a relatively simple question-answering test, only obscured by the temporal reframing.  Because the test year was 1940, the information needed to answer the questions correctly should be present in the training material at the cut-off dates for all models.  We do acknowledge that there are mistakes both in the corpus and the grader precluding perfect scores.  We did, however, look into the mistakes, and found that they are not of a scale that could not explain all the mistakes that the models made.

We also note that the NorwAI-Magistral24B model, which is the largest publicly available model for Norwegian―with reasoning capabilities―performs well on the task. It is within striking distance of international models of similar size. Because we only compared to two other Norwegian models, so more experiments are needed to conclude that it is the only Norwegian-focused model that can do well on this TR task, but it cannot rule it out.  On a similar note, it could also be the chain-of-thought trace offered by the Magistral architecture that facilitated the good performance.  It is also worth noting that the effect of using warm (on top of pre-trained weights) models had a large impact on this task vs. the cold models (built from scratch).

We also note that gpt-oss was used as the evaluator, its scores could have been inflated when it was also doing the grading.  We were simply curious about how it would do when it was "marking its own papers", if it could somehow tell that the answers it graded came from querying the same model.  While interpret these results with caution (no manual grading or alternative LLM-as-judge were used), we present the results as the experiments were run.  The results do not seem unreasonable, however, given the reported performance of gpt-oss on other tasks. 

We also note that while The Book is an obscure Norwegian trivia book from 1940, it is publicly available, and we cannot rule out that it has formed, or will form, as part of the training material for either pre- or post-training of LLMs.  We have, however, not seen any indication that this has been the case.

The Llama3.1:405B showed better performance for the Norwegian prompting, as opposed to the other models in Group A.  It is not clear to us why this shift occurred.  We also experimented with both the 70B and 70B-Instruct models in this family, but saw little difference.  The same score for English prompting and slightly worse for the instruct-tuned version for Norwegian.

Finally, we note that our prompts could have been phrased more precisely, as one question had an answer from the winter from March 1940, which means that the cut-off date should have been set to that month at the earliest.  We do not, however, consider that this imprecision affected our results.

\section{Conclusions and Future Work}

In this work, we looked at two main Research Questions.  First, the effect of scaling on Temporal Reasoning, and next, the effect of prompting in English vs. Norwegian.

\textbf{RQ1}: As expected, we found that performance on the task scales with parameter size for all LLM families we experimented with.  The mid-eighties top performance suggests that there is room for improvement.

\textbf{RQ2}: We saw that prompting in English (but answering in Norwegian) improved the performance of most models with the exception of the largest Llama models. 

\subsection{Future Work}
We would like to expand on this work by further analyzing the effects of different levels of quantization on Temporal Reasoning.  The present work only used one quantization size for each model, and knowing the effects of increasing the bits per weight (as well as no quantization) would be interesting.  It is reasonable to assume better performance, but the magnitude is unclear.

Because of the encouraging results on the NorwAI-Magistral24B model, we would like to investigate further whether it came as a consequence of increased parameter size or the reasoning capabilities.
 
We would also like to expand on this work on TR by using other old trivia books in other non-English languages.  It is also possible that the LLMs would do better on this task by means of Retrieval-Augmented Generation (RAG) solutions that could, for instance retrieve articles on the history of countries.

Finally, we would also like to experiment with creating instruction-data from books of this type (as long as the copyright is expired), either directly or synthetically, e.g., by multiplying them with paraphrasing, to experiment with the effect on other TR tasks and datasets.

\section{Acknowledgments}

The computations were performed on resources provided by Sigma2 - the National Infrastructure for
High-Performance Computing and Data Storage in Norway.

We would also like to thank Dhruv Gupta for valuable feedback on the manuscript.

\section{Limitations}

Because The Book is under copyright protection, the corpus we used for querying the LLMs cannot be published without permission, as this would be equivalent to sharing the book.  However, we consider presenting the results and quoting from the book to be fair use.   The book is publicly available in Norwegian libraries. All translations into English are done by the authors.

\section{Ethics Statement}

Since this book was published in 1940, when Norway was under Nazi occupation, The Book had passed through the censorship regime in place.  As a consequence, some questions could potentially contain offensive wording or other problematic bias.  However, The Book was designed as a trivia book with the purpose of testing easily looked-up tidbits of information. A native speaker examined the questions and found four questions using the Norwegian term for Jews, of which none contained derogatory wording. (An example translates to \emph{"What does the first part of the Old Testament contain, which according to Jewish partitioning is called 'The Law'?"}) As a consequence, they were kept in the presented scores.  Three questions referenced the Norwegian term for "negro", but were not derogatory outside the use of the term itself. Because the term is considered offensive in 2025, however, we removed the questions from the scores. The stronger models answered the questions correctly (and without objection), which would have increased their scores by ~3‰.

Furthermore, we acknowledge that we do not fully know how the models we have used in these experiments are trained, especially with regards to instruction-data.  However, as long as we are able to download the model checkpoints and do inference on them locally, we find their use acceptable also in a scientific context.  They provide good information on the progress of models that have been developed with full transparency.

\section{Bibliographical References}\label{sec:reference}

\bibliographystyle{lrec2026-natbib}
\bibliography{vetdedet}


\appendix
\section{Appendix}
\label{appendix}
\subsection{Prompts Used}
\textbf{System Prompt:}
All evaluated models were prompted with this system prompt in English or Norwegian before receiving
questions from the dataset:
\begin{quote}
\itshape
You are an expert in answering history quizzes using only knowledge from up to 1940. Answer in one
or two sentences, using only knowledge up to 1940.'' \end{quote} \begin{quote} \itshape Du er
ekspert på norsk språk og verdenshistorie før 1940. Svar kort (1--2 setninger) og bruk kun kunnskap
før 1940. Svar på norsk''
\end{quote}
\textbf{Grader Prompt:}
The grader was initialized with this system prompt before being presented with the LLM agent answer
and the correct answer from the book:
\begin{quote}
\itshape
``You are an expert in grading history quizzes using only knowledge from up to 1940. Given a
question, a test answer, and the correct answer, output T if the test answer is essentially correct,
or F if it is wrong. OUTPUT EXACTLY 1 CHARACTER AND NOTHING ELSE. example: question: What's the
capital of Egypt? Model Answer: Paris Correct answer: Cairo Your output: F''
\end{quote}
\end{document}